\documentclass[5p]{elsarticle}
\pdfoutput=1
\usepackage{graphicx}
\usepackage{amsmath, amsthm, amssymb}

\usepackage[english]{babel}

\newcommand{\vectornorm}[1]{\left|\left|#1\right|\right|}

\begin{document}


\title{Readouts for Echo-State Networks Built using Locally Regularized Orthogonal Forward Regression}

\author[kyushu]{J\'an Dolinsk\'y\corref{cor1}}
\ead{jan.dolinsky@ymail.com}

\author[kyushu]{Kei Hirose}
\ead{mail@keihirose.com
}
\author[kyushu]{Sadanori Konishi}
\ead{konishi@math.chuo-u.ac.jp}

\address[kyushu]{Department of Mathematics, Kyushu University, 744 Motooka, Nishi-ku, Fukuoka, Japan}

\cortext[cor1]{Corresponding author}



\begin{abstract}
Echo state network (ESN) is viewed as a temporal non-orthogonal expansion with pseudo-random parameters.
Such expansions naturally give rise to regressors of various relevance to a teacher output.
We illustrate that often only a certain amount of the generated echo-regressors effectively explain the variance of the teacher output
and also that sole local regularization is not able to provide in-depth information
concerning the importance of the generated regressors.   
The importance is therefore determined by a joint calculation of the individual variance contributions and Bayesian relevance 
using locally regularized orthogonal forward regression (LROFR) algorithm.
This information can be advantageously used in a variety of ways for an in-depth analysis of an ESN structure and its state-space parameters
in relation to the unknown dynamics of the underlying problem.
We present locally regularized linear readout built using LROFR.
The readout may have a different dimensionality than an ESN model itself, and besides improving robustness and accuracy of an ESN
it relates the echo-regressors to different features of the training data and may determine what
type of an additional readout is suitable for a task at hand.
Moreover, as flexibility of the linear readout has limitations and might sometimes be insufficient for certain tasks,
we also present a radial basis function (RBF) readout built using LROFR. It is a flexible and parsimonious readout
with excellent generalization abilities and is a viable alternative to readouts based on a feed-forward neural network (FFNN)
or an RBF net built using relevance vector machine (RVM).
\end{abstract}


\maketitle

\section{Introduction}
\label{sec:I}
ESNs are a novel class of recurrent neural networks (RNN) \cite{Jaeger2001}. 
Their easy construction and simple training procedure are appealing and
have attracted the attention of many researchers.
The ESN model consists of the state-space update equation (\ref{eq:esnStateSpace}) and the readout equation (\ref{eq:esnReadout})
\begin{equation}\label{eq:esnStateSpace} 
\mathbf{x}(k+1) = \mathbf{f}(\mathbf{W}^{in}\mathbf{u}(k+1) + \mathbf{W}\mathbf{x}(k) + \mathbf{W}^{fb}\mathbf{y}(k))
\end{equation}
\begin{equation}\label{eq:esnReadout}
\mathbf{y}(k+1) = \mathbf{W}^{out}[\mathbf{u}(k+1), \mathbf{x}(k+1)]
\end{equation}
where $\mathbf{u}(k)$ is an $L$-dimensional input vector, $\mathbf{y}(k)$ is an $P$-dimensional output vector and 
$\mathbf{x}(k)$ is an $M$-dimensional echo-state vector. $\mathbf{W}^{in} \in \mathcal{R}^{M \times L}$ denotes an input weight matrix,
$\mathbf{W} \in \mathcal{R}^{M \times M}$ denotes an internal weight matrix and $\mathbf{W}^{fb} \in \mathcal{R}^{M \times P}$
is a feedback matrix. Vector function $\mathbf{f}$ is applied element-wise to its arguments.
The most common choice for $\mathbf{f}$ is either a vector of sigmoid or identity functions. 

Mathematically, the state-space equation (\ref{eq:esnStateSpace}) represents a non-orthogonal temporal expansion
of teacher and input signal onto a higher-dimensional space.
The expansion is carried out so that diverse echoes of an input and teacher signal are generated (hence the name echo-state).  
This diversity, which should appropriately "explain" a variance of a teacher signal, is the key to the successful training of an ESN.
Traditionally, the weight matrices $\mathbf{W}^{in}$, $\mathbf{W}$, $\mathbf{W}^{fb}$
are generated in a pseudo-random manner with no complicated estimation being involved \cite{Jaeger2001}.
Nonlinearities in $\mathbf{f}$ are usually chosen by a trial-and-error approach \cite{Jaeger2001, Jaeger2002}.
This approach usually suffices to generate many diverse signals.
Signals generated by the expansion are successively used as regressors by the linear regression readout mechanism 
with the readout weights $\mathbf{W}^{out}$ being the only parameters to be estimated.

Although the training procedure is simple and transparent, constructing an ESN model that generalizes well
is not straightforward and usually involves a considerable number of trials with different ESN parameters.
There has been a considerable effort to improve the original ESN model.
Some optimizations can be done before expanding echo-signals in state-space parameters
(mostly in the weight matrices and in the vector of update functions $\mathbf{f}$)
\cite{Ishii2004, Babinec2006joei, Deng2007, Steil2007, Dolinsky2010} and some improvements are possible after the expansion
in the readout equation \cite{Babinec2006icann, Holzmann2010, Ozturk2007, Dutoit2009, ShuEtal2012}.

The motivation for adjusting the various parameters of the state-space equation is to obtain a stable state-space update 
and to generate echo-regressors so that their linear compounds explain the variance of the response variable sufficiently
well - up to a desired accuracy.
Stability of the state-space update is of crucial importance but research along this line is not targeted here.
This study analyzes the statistical quality of regressors generated by the ESN expansion 
and shows how results of such analysis can be advantageously used in a variety of ways as described later in the text. 

The state-space equation generates non-orthogonal echo-regressors of various relevance to the response variable.
Some of the generated echo-regressors explain a more significant part of the response variable variance and some less.
The number of echo-regressors is often excessively high (up to few hundreds or more) with
some of the generated echo-regressors often being collinear (multicollinearity). 
A high number of echo-regressors and multicollinearity contributes to the undesired numerical instability of the linear readout regression model
where a small change in regressors causes a large change in the response.   
Generalization abilities of such an unstable model are usually low and the model itself might be meaningless. 
Moreover, models with many parameters may easily fit into the noise of the response variable and are thus prone to overfitting.
From the standpoint of regression modeling, it is therefore desirable to reduce the number of readout parameters, especially
by removing (or penalizing) collinear and less significant regressors. 
It is also worth to point out that echo-regressors are "loosely" coupled only in the state-space equation.
The readout mechanism may therefore use selected significant echo-regressors only so that a stable readout model
with a good generalization performance is obtained.
A traditional linear regression readout measures the suitability of fit only in terms of mean square error (MSE).
This is often unsatisfactory because no information concerning the quality of the generated echo-regressors is given.

Various improvements addressing the mentioned issues have been proposed \cite{Dutoit2009, ShuEtal2012}.
Ridge regression combined with pruning of output weights was tested in \cite{Dutoit2009}.
Pruning decreased the undesired effects of multicollinearity and this resulted in an increased performance and better generalization.
Pruning is however computationally exhaustive and ridge regression regularizes a least-squares (LS) solution only globally
using a single regularization parameter which is estimated using the grid search.    
Performance was observed only in terms of MSE which limits further analysis.
Work in \cite{ShuEtal2012} presents a joint estimation of
local (individual) regularization parameters and delay parameters in delay\&sum (D\&S) readout based on a variational bayesian approach.
The algorithm is computationally effective and capable of an accurate estimation of delay parameters 
which vastly improves the memory capacity of an ESN and provides a deeper insight into the temporal structure of the underlying problem.
Joint estimation of regularization parameters then further enhances the robustness of the model.

Regularization parameters smooth model response by penalizing individual echo-regressors
according to their "relevance" to the smoothed "noise-free" version of the teacher signal.
This alone while significantly improving generalization does not fully indicate the usefulness of the individual echo-regressors. 
Our work experimentally illustrates this phenomenon  
and we propose to study individual echo-regressors using locally regularized orthogonal forward regression
(LROFR) so that both their individual variance contributions and their Bayesian relevance is jointly determined.
This can be advantageously used in a variety of ways. 
The appropriate dimensionality for an ESN readout may be determined and
the readout may have a lower dimensionality than the state-space update itself which naturally boosts the robustness
and is useful when e.g. augmenting echo-states.
The effectiveness of an expansion is transparently determined and this enables an in-depth evaluation of ESN state-space parameters
and ESN structure; e.g. determining the suitability of an additional nonlinear readout discussed shortly or
other aspects like analysis of multiple reservoirs etc. discussed later in the text.

A linear readout in general has, however, limitations in terms of the model's flexibility which might sometimes be insufficient for a task at hand.
LROFR analysis is able to determine whether this is the case and suggests whether a more flexible readout mechanism should be used.
Such a readout usually requires less parameters than a linear readout and approximates the response variable with better accuracy.
Feed forward neural networks (FFNN) and RBF nets constructed using Relevance Vector Machine (RBF-RVM)
are perhaps the most popular examples of such models \cite{KroseSmagt96, Tipping2001, Chen2006}.
Advantages of the both models is that they often reduce the number of readout parameters, attenuate effects of multicollinearity by using
entire internal state as an input, and often explain a higher portion of the variance of the response variable than linear compounds of original
echo-regressors in a linear readout \cite{Babinec2006icann}.
Training of an FFNN requires a nonlinear least-squares iterative search algorithm, based on gradients, which has a considerably high computational cost.
The extreme flexibility of this model requires cautious training to prevent over-fitting and obtaining a meaningful model is not straightforward.
This is contrary to the fast and appealing LS estimation of parameters in a traditional linear readout of an ESN.
Kernel support vector machines such as RBF-RVM received considerable attention and popularity over the last decade
and may also be used as a readout mechanism for an ESN.
In comparison to FFNN, RBF-RVM is easier to train and possesses some computational advantages.
Recently, however, it was shown that RBF-RVM suffers numerical instability when dealing with complex and highly nonlinear data
and that its abilities have perhaps been overstated \cite{Chen2006}.

Locally regularized RBF readout where RBF net is constructed using LROFR with D-Optimality cost (RBF-LROFR-DOPT)
is presented for cases where LROFR analysis indicates that a more flexible readout should be used.
This readout has an excellent generalization performance and possesses computational advantages to readouts based on FFNN/RBF-RVM \cite{Chen2006}.

The paper is organized as follows. Variable selection that is used to analyze the individual echo-regressors is briefly presented first.
Next, locally regularized RBF models are explained in Section~\ref{sec:RBF}. 
Analysis of an ESN using the presented variable selection mechanisms is then provided in Section~\ref{sec:III}
and results in a proposal of locally regularized linear readout. Locally regularized RBF readout is subsequently presented as a viable alternative
to readouts based on FFNN/RBF-RVM. Section~\ref{sec:modelingExamples} illustrates our modeling strategy using
a real-world noisy task and a synthetic noiseless task.
Future directions for research are suggested in the discussion of Section~\ref{sec:discussion}.
Finally, the merits of the proposed modeling strategy are summarized and new prospective applications are outlined in the conclusion.   


\section{Variable Selection}
\label{sec:OFR}
\subsection{Orthogonal Forward Regression}
The Orthogonal Forward Regression (OFR) algorithm uses the advantages of orthogonal decomposition
of the design matrix to compute the individual contribution of each regressor to the response variable variance \cite{Chen1989}.

The regression model is to be built from N-sample data set $\{\mathbf{x}(k),y(k)\}_{k=1}^{N}$
where $\mathbf{x}(k) \in \mathcal{R}^{M}$ and $y(k) \in \mathcal{R}$ are the $k$-th training input vector
and corresponding desired scalar response, respectively.
The regression model may be expressed in the matrix notation
\begin{equation}
\mathbf{y = X\boldsymbol{\beta} + e}
\end{equation}
where $\mathbf{y} = [y(1),\dots,y(N)]^{T}$, $\mathbf{e} = [e(1),\dots,e(N)]^{T}$, $\boldsymbol{\beta} = [\beta_{1},\dots,\beta_{M}]^{T}$,
and $\mathbf{X}$ is the design matrix of size $N\times M$ with its rows $\mathbf{x}(1),\dots,\mathbf{x}(N)$.
Orthogonal decomposition of the design matrix $\mathbf{X}$ may be written as follows
\begin{equation} \label{eq:ols1}
\mathbf{X = QR}
\end{equation}
where
\begin{equation} \label{eq:ols1a}
\mathbf{Q} = [\mathbf{q}_{1}, \dots, \mathbf{q}_{M}]
\end{equation}
is a design matrix with orthogonal columns $\mathbf{q}_{i}$ satisfying $\mathbf{q}_{i}^{T}\mathbf{q}_{j} = 0$ if $i \neq j$
and $\mathbf{R}$ is an upper triangular matrix.
\begin{equation} \label{eq:ols1b}
\mathbf{R} = 
\begin{bmatrix}
  1 & r_{1,2} & \cdots & r_{1,M} \\
  0 & 1 & \cdots & \vdots \\
  \vdots  & \vdots  & \ddots & r_{M-1,M}  \\
  0 & \cdots & 0 & 1
\end{bmatrix}
\end{equation}
The regression model can be then expressed using the orthogonal design matrix $\mathbf{Q}$.
\begin{equation} \label{eq:ols2}
\mathbf{y = Qg + e}
\end{equation}
Minimizing standard least-squares error criterion $J=\mathbf{e}^{T}\mathbf{e}$ (i.e. setting @J/@g=0)
yields a vector of regression coefficients $\mathbf{g} = (\mathbf{Q}^{T}\mathbf{Q})^{-1}\mathbf{Q}^{T}\mathbf{y}$
which satisfies the triangular system $\mathbf{R\boldsymbol{\beta} = g}$.
Solving the triangular system obtains the original regression coefficients $\boldsymbol{\beta}$.
Although computing the regression coefficients in such a way has advantages \cite{Wong1935},
the algorithm here is concerned with a \emph{variable selection} in a forward regression manner using 
the advantages of orthogonality.
 
Some useful transformations may be carried out with the orthogonal regression model as follows.
\begin{equation} \label{eq:ols3}
\mathbf{y} = g_{1}\mathbf{q}_{1} + \dots + g_{M}\mathbf{q}_{M} + \mathbf{e}
\end{equation}
Orthogonality of the regressors $\mathbf{q}_{i}$ allows us to write
\begin{equation} \label{eq:ofr4}
\mathbf{y}^{T}\mathbf{y} = g_{1}^{2}\mathbf{q}_{1}^{T}\mathbf{q}_{1} + \dots + g_{M}^{2}\mathbf{q}_{M}^{T}\mathbf{q}_{M} + \mathbf{e}^{T}\mathbf{e}
\end{equation}
If mean of the response variable $\mathbf{y}$ is $0$ then its variance equals to
\begin{equation} \label{eq:ols5}
N^{-1}\mathbf{y}^{T}\mathbf{y} = N^{-1}(g_{1}^{2}\mathbf{q}_{1}^{T}\mathbf{q}_{1} + \dots + g_{M}^{2}\mathbf{q}_{M}^{T}\mathbf{q}_{M}) + N^{-1}\mathbf{e}^{T}\mathbf{e}
\end{equation}
The variance of $\mathbf{y}$ is therefore expressed by the variance explained by the model (the regressors) and unexplained variance
of the error term. 
Because the regressors do not interact (they are orthogonal), it is possible to compute their 
individual variance contributions to $\mathbf{y}$.
Each individual regressor increases the variance explained by the model and 
reduces the unexplained variance of the error term.
This error reduction ratio for an $i$-th single regressor can be expressed as follows.
\begin{equation} \label{eq:ols6}
err_{i} = N^{-1}g_{i}^{2}\mathbf{q}_{i}^{T}\mathbf{q}_{i} / N^{-1}\mathbf{y}^{T}\mathbf{y} 
= g_{i}^{2}\mathbf{q}_{i}^{T}\mathbf{q}_{i} / \mathbf{y}^{T}\mathbf{y}
\end{equation}
The variance equation~(\ref{eq:ols5}) may be then alternatively expressed as follows
\begin{equation} \label{eq:ols7}
1 = \sum_{i=1}^{M} err_{i} + \mathbf{e}^{T}\mathbf{e}/\mathbf{y}^{T}\mathbf{y}
\end{equation}
and the unexplained variance ratio then simply is
\begin{equation} \label{eq:ofr7b}
\mathbf{e}^{T}\mathbf{e}/\mathbf{y}^{T}\mathbf{y} = 1 - \sum_{i=1}^{M} err_{i}.
\end{equation}
The algorithm builds a sub-model by selecting $M_{sub}$ significant regressors (usually $M_{sub} \ll M$)
 in a \emph{forward regression} manner based on the error reduction ratio.
Selection is terminated when user-specified tolerance $0 < \xi < 1$ for the unexplained variance ratio is reached.
\begin{equation} \label{eq:ols8}
1 - \sum_{i = 1}^{M_{sub}} err_{i} < \xi
\end{equation}
Alternatively, all available regressors are gradually selected and the unexplained variance ratio 
is observed after each selection. It is often the case that after selecting a certain number of 
significant regressors, introducing more regressors causes the unexplained variance ratio
to decrease only marginally. Such regressors often contribute to ill-conditioning 
and the corrupt overall statistical quality of the model.

\subsection{Locally Regularized OFR}
Regression models built using OFR may still fit into the noise of the response (overfitting) because
selection is based purely on minimization of error.
Local regularization appropriately smoothes model response by penalizing regressor terms to prevent overfitting.
Locally regularized OFR (LROFR) \cite{Chen2003, Chen2006} adopts the following regularized error criterion
\begin{equation} \label{eq:lrofr1}
J_{R}(\mathbf{g},\boldsymbol{\lambda}) = \mathbf{e}^{T}\mathbf{e} + \sum_{i=1}^{M} \lambda_{i}g_{i}^{2}
= \mathbf{e}^{T}\mathbf{e} + \mathbf{g}^{T}\mathbf{\Lambda}\mathbf{g}
\end{equation}
where $\mathbf{\Lambda} = diag\{\lambda_{1},\dots,\lambda_{M}\}$ and $\lambda_{i}$ is an $i$-th regularization parameter.
As in the case of the OFR, Eq.~(\ref{eq:ofr4}), it can be shown \cite{Chen2003} that the orthogonal regression model may be written as
\begin{equation} \label{eq:lrofr2}
\mathbf{y}^{T}\mathbf{y} = \sum_{i-1}^{M}g_{i}^{2}(\mathbf{q}_{i}^{T}\mathbf{q}_{i} + \lambda_{i})+ \mathbf{e}^{T}\mathbf{e} + \mathbf{g}^{T}\mathbf{\Lambda}\mathbf{g}
\end{equation}
and the error criterion can be thus expressed as
\begin{equation} \label{eq:lrofr3}
\mathbf{e}^{T}\mathbf{e} + \mathbf{g}^{T}\mathbf{\Lambda}\mathbf{g} = \mathbf{y}^{T}\mathbf{y} - \sum_{i-1}^{M}g_{i}^{2}(\mathbf{q}_{i}^{T}\mathbf{q}_{i} + \lambda_{i}).
\end{equation}
Similarly to the Eq.~(\ref{eq:ofr7b}), normalizing (\ref{eq:lrofr3}) by $\mathbf{y}^{T}\mathbf{y}$ gives
\begin{equation} \label{eq:lrofr4}
(\mathbf{e}^{T}\mathbf{e} + \mathbf{g}^{T}\mathbf{\Lambda}\mathbf{g})/\mathbf{y}^{T}\mathbf{y} = 1 - \sum_{i-1}^{M}g_{i}^{2}(\mathbf{q}_{i}^{T}\mathbf{q}_{i} + \lambda_{i})/\mathbf{y}^{T}\mathbf{y}.
\end{equation}
Analogously to the OFR, the regularized error reduction ratio due to $\mathbf{q}_{i}$ is then
\begin{equation} \label{eq:lrofr5}
rerr_{i} = g_{i}^{2}(\mathbf{q}_{i}^{T}\mathbf{q}_{i} + \lambda_{i}) / \mathbf{y}^{T}\mathbf{y}.
\end{equation}
Significant regressors are selected based on $rerr$ criterion in a forward regression manner \cite{Chen2003}.
Selection finishes when user-specified tolerance $0 < \xi < 1$ is reached.
\begin{equation} \label{eq:LROFR6}
1 - \sum_{i = 1}^{M_{sub}} rerr_{i} < \xi
\end{equation}
This produces a sparse model of $M_{sub} \ll M$ regressors. 

Regularization parameters $\lambda_{i}$ are initially unknown. They are all set to a small value (e.g. 0.01) and a pass of an OFR using $rerr$
criterion is carried out over the initial full model. 
$\lambda_{i}$ of a resulting sub-model is updated using the Bayesian evidence procedure \cite{Chen2003} and
an OFR using $rerr$ criterion with the updated $\lambda_{i}$ is carried out over the previously generated sub-model.
This procedure is applied iteratively until $\lambda_{i}$ remains sufficiently unchanged between two iterations.
In general, each iteration generates a sub-model from a model generated in the previous iteration and updates $\lambda_{i}$ used in a next iteration. 
This can be schematically expressed as follows.
\begin{equation} \label{eq:LROFR7}
\mathbf{X}^{full} \rightarrow \mathbf{X}^{(2)} \rightarrow \dots \rightarrow \mathbf{X}^{final}
\end{equation}
where $M_{full} \geq M_{2} \geq \dots \geq M_{final}$.
The algorithm is computationally effective and the number of regressors often decreases dramatically within the first few (e.g. 4-5) iterations.
A few more iterations are usually required for regularization parameters to converge.
Typically about 10 iterations in total suffice to construct final parsimonious model with the design matrix $\mathbf{X}^{final}$. 

\subsection{Locally Regularized OFR with D-Optimality Cost}
LROFR algorithm can be further enhanced by the D-Optimality cost which effectively maximizes determinant of $(\mathbf{X}^{T}\mathbf{X})$
and thus improves model robustness \cite{Chen2003}. LROFR with D-Optimality cost adopts the combined criterion
\begin{equation} \label{eq:LROFR_DOPT1}
J_{CR}(\mathbf{g},\boldsymbol{\lambda},\beta) = J_{R}(\mathbf{g},\boldsymbol{\lambda}) + \beta \sum_{i=1}^{M}-log(\mathbf{q}_{i}^{T}\mathbf{q}_{i}).
\end{equation}
The algorithm is identical to the LROFR but the selection is governed by the combined regularized error reduction ratio.
\begin{equation} \label{eq:LROFR_DOPT2}
crerr_{i} = g_{i}^{2}((\mathbf{q}_{i}^{T}\mathbf{q}_{i} + \lambda_{i}) + \beta log(\mathbf{q}_{i}^{T}\mathbf{q}_{i}))  / \mathbf{y}^{T}\mathbf{y}
\end{equation}
Stopping rule (\ref{eq:LROFR6}) for a single iteration is not necessary any more because after selecting a certain number of 
significant regressors all remaining unselected regressors will have their $crerr_{i} \leq 0$.
Finding an appropriate value of the user-specified parameter $\beta$ is usually quick and straightforward.


\section{RBF Regression Modeling}
\label{sec:RBF}
The aim in nonlinear system identification is to approximate (identify) dynamics between observed inputs and outputs of interest.
The discrete-time system to be identified is in the following form
\begin{equation}
\label{eq:RBF1} 
y(k) = f(\mathbf{u}(k),...,\mathbf{u}(n-l_{u});y(k-1),...,y(k-l_{y})) + e(k); 
\end{equation}
where $\mathbf{u}(k)$ is the input vector and $y(k)$ is the 
scalar\footnote{Extension to multi-dimensional vector output is straightforward, scalar notation is used for sake of simplicity.}
output for the time step $k$,
$l_{u}$ and $l_{y}$ are the time delay lags in the input and output respectively and $e(k)$ is the system white noise.
Letting 
\begin{equation} \label{eq:RBF2} 
\mathbf{x}(k) = [\mathbf{u}(k),...,\mathbf{u}(n-l_{u});y(k-1),...,y(k-l_{y})]
\end{equation} 
reduces~(\ref{eq:RBF1}) into
\begin{equation} \label{eq:RBF3} 
y(k) = f(\mathbf{x}(k)) + e(k); 
\end{equation} 
The system is to be identified from N-sample data set $\{\mathbf{x}(k),y(k)\}_{k=1}^{N}$.  
One of the possible approaches to approximate $f$ is the RBF regression modeling.
\begin{equation} \label{eq:RBF4} 
y(k) = \hat{y}(k) + e(k) = \sum_{i=1}^M \theta_{i} \phi_{i}(\mathbf{x}(k)) + e(k),
\end{equation} 
where $M$ denotes the number of RBF regressors, $\theta_{i}$ is the i-th regression coefficient and
$\phi_{i}(\mathbf{x}(k))$ is its corresponding i-th RBF regressor with centre $\mathbf{c}_i$ and variance $\upsilon_{i}$.
\begin{equation} \label{eq:RBF5}
\phi_{i}(\mathbf{x}(k)) = \varphi(\vectornorm{\mathbf{x}(k) - \mathbf{c}_{i}}/\upsilon_{i}) 
\end{equation} 
$\vectornorm{\mathbf{.}}$ is the Euclidean norm and function $\varphi$ is an RBF nonlinearity.
The thin plate spline (TPS) function (Eq.~\ref{eq:RBF6}) and the Gaussian function (Eq.~\ref{eq:RBF7}) are the two most common choices.
\begin{equation} \label{eq:RBF6}
\varphi(\chi/1) = \chi^{2}log(\chi) 
\end{equation} 
\begin{equation} \label{eq:RBF7}
\varphi(\chi/\upsilon) = e^{-\chi^{2}/\upsilon^{2}} 
\end{equation}

The regression model of Eq.~(\ref{eq:RBF4}) can be written in a matrix form
\begin{equation} \label{eq:rbfMatrixForm}
\mathbf{y = \Phi\boldsymbol{\theta} + e}
\end{equation}
where $\mathbf{y} = [y(1),\dots,y(N)]^{T}$, $\mathbf{e} = [e(1),\dots,e(N)]^{T}$, $\boldsymbol{\theta} = [\theta_{1},\dots,\theta_{M}]^{T}$,
and $\mathbf{\Phi} = [\boldsymbol{\phi}_{1},\dots,\boldsymbol{\phi}_{M}]^{T}$ is the design matrix with its columns
$\boldsymbol{\phi}_{i} = [\phi_{i}(\mathbf{x}(1)),\dots,\phi_{i}(\mathbf{x}(N))], 1 \leq i \leq M$.

The crucial issue in the RBF modeling is to choose a set of basis functions $\phi_{i}$ so that
resulting model is robust, parsimonious and generalizes well.
A common approach is to generate many basis functions first, and then iteratively select a suitable subset until a final parsimonious model is generated.
Each generated $\phi_{i}$ is defined by its centre $\mathbf{c}_{i}$ and variance $\upsilon_{i}$.
The variance $\upsilon_{i}$ is often fixed to a single value $\upsilon$ for each $\phi_{i}$.
Let $\mathbf{c}_{i} = \mathbf{x}(i), 1\leq i \leq M$, and set $M = N$ which effectively makes every data point $\mathbf{x}(i)$ a candidate for a centre.
This makes the number of generated basis functions $\phi_{i}$ equal to the number of data points and
matrix $\mathbf{\Phi}$ will result in size of $N\times N$.
Let the design matrix of this initial full model be denoted by $\mathbf{\Phi}^{full}$ and
the design matrix of the final model by $\mathbf{\Phi}^{final}$. 
LROFR algorithm combined with D-Optimality cost is used to 
iteratively select useful regressors from $\mathbf{\Phi}^{full}$ ($N\times M_{full}$) into 
$\mathbf{\Phi}^{final}$ ($N\times M_{final}$) where $M_{final} << M_{full}$ \cite{Chen2003, Chen2006}.
In each iteration, the algorithm selects a subset of regressors from the previous iteration in a forward regression manner
and updates the regularization parameters used in the next iteration.


\section{ESN Regression Modeling} 

\label{sec:III}
\subsection{Linear Readout}
The state-space update equation~(\ref{eq:esnStateSpace}) is used to extract temporal or spatial patterns from original
N-sample data set $\{\mathbf{u}(k),\mathbf{y}(k)\}_{k=1}^{N}$
where $\mathbf{u}(k) \in \mathcal{R}^{L}$ and $y(k) \in \mathcal{R}^{Q}$ are the $k$-th training input vector
and corresponding desired vector response, respectively.
Echo-states $\mathbf{x}(k+1)$ are sampled via the Eq.~(\ref{eq:esnStateSpace}) using the original input vectors $\mathbf{u}(k+1)$
and/or output vectors $\mathbf{y}(k)$, and previous echo-state $\mathbf{x}(k)$ itself.  
Some initial echo-states from the sampling are discarded to avoid the influence of the undefined random states
$\mathbf{x}(0)$ and $\mathbf{y}(0)$ at time $k = 0$.
The rest of the echo-states are then stored as rows of the design matrix $\mathbf{X}$.
Let rows of the matrix $\mathbf{y}$ be the vectors $\mathbf{y}(k)^{T}$. 
The ESN regression readout model can be then expressed in the matrix form
\begin{equation} \label{eq:esn3}
\mathbf{y = X}\mathbf{W}^{out} + \mathbf{e}
\end{equation}
where $\mathbf{W}^{out}$ are the regression weights (coefficients) and $\mathbf{e}$ are random errors with 
common variance and zero mean. 
The regression weights $\mathbf{W}^{out}$ are traditionally estimated using the least-squares (LS) method
\begin{equation} \label{eq:esn4}
\mathbf{W}^{out} = (\mathbf{X}^{T}\mathbf{X})^{-1}\mathbf{X}^{T}\mathbf{y}.
\end{equation}

Sampling via the state-space Eq.~(\ref{eq:esnStateSpace}) generates echo-regressors (columns of the design matrix $\mathbf{X}$)
of various relevance to the desired response. The shape and characteristics of the echo-regressors are indirectly governed
by the ESN state-space parameters
(pseudo-random weight matrices $\mathbf{W}^{in}$, $\mathbf{W}$, $\mathbf{W}^{fb}$, nonlinearity $\mathbf{f}$, etc.).
These parameters are adjusted (mostly using intuition and experience) so that the linear compounds of the generated echo-regressors
approximate the response with a desired accuracy.

Obviously, an expansion with pseudo-random parameters gives rise to regressors of various importance.
As outlined in the introduction, the number of echo-regressors is often excessively
high with some of the echo-regressors being collinear. High dimensionality of the readout and multicollinearity contributes
to numerical instability of the readout and it is often the case that LS estimates of $\mathbf{W}^{out}$ have large variances and inappropriately
large mean values. Models with many parameters are also prone to over-fitting because they may easily fit into the noise of the response.
Taking these issues into account, it is desirable to reduce dimensionality of the readout, particularly by removing or penalizing the undesired
collinear and less significant echo-regressors.

These issues may be addressed by local regularization applied to all the generated echo-regressors \cite{ShuEtal2012}. 
Regularization penalizes individual echo-regressors based on their Bayesian relevance so that the readout response is smooth and does not fit the into noise.  
This alone while significantly improving generalization does not fully indicate the importance of the individual echo-regressors. 
Examination of the regularization parameters will only show which regressors had to be attenuated so that model response is "smooth enough".
Such analysis is therefore unsatisfactory for relating particular echo-regressors and overall ESN structure to different features of the training data.

To overcome this limitation we study the individual quality of the echo-regressors using LROFR. 
Joint analysis of the individual variance contributions and Bayesian relevance of the echo-regressors using LROFR is able
to give more accurate information concerning the importance of particular regressors.
This information can be used in a variety of ways for an in-depth analysis of an ESN structure and its state-space parameters
in a relation to the unknown dynamics of the underlying problem.


In the following we will describe how to compute the individual significance of echo-regressors by using LROFR algorithm.  

\subsection{Analysis of echo-regressors using the OFR algorithm}

The individual variance contributions of the echo-regressors can be calculated using the OFR algorithm given in Section~\ref{sec:OFR}.
The stopping rule~(\ref{eq:ols8}) is left out and all generated echo-regressors are gradually selected
into the readout model based on their respective error reduction ratios.
The unexplained variance ratio (variance ratio of the error term)
is observed after each selection to see by how much an introduction of a new echo-regressor decreases
the ratio. 
Some of the generated echo-regressors explain a higher portion of the response variance and some less.
It is therefore often the case that in the beginning of the selection process the unexplained variance ratio decreases sharply
while selecting more significant regressors into the readout model.
Gradually, the unexplained variance ratio starts to decrease substantially slower when the algorithm selects from
the remaining echo-regressors that are either non-significant or collinear with any of the already selected echo-regressors.
Observing this selection process reveals the redundancy of an ESN expansion; i.e. roughly how many of the echo-regressors are important
and how many are not so. 
This information may be used to determine the appropriate dimensionality for the linear readout so that a parsimonious and more stable
readout is obtained.
Moreover, it can in turn be used to suggest the dimensionality of the state-space equation and it helps to evaluate suitability
of its pseudo-random parameters too.

This analysis is also able to determine whether more flexibility is required
either in readout response surface or in readout temporal processing capability.
If the analysis indicates a need for a more flexible response surface, one may use
RBF readout built using LROFR-DOPT which is presented later. 
D\&S readout may be used in cases where a need for more temporal capability is indicated. 

\subsection{Locally regularized linear readout}
Analysis using OFR shows how many and which of the generated echo-regressors are useful for modeling a task of interest. 
The principle of selecting echo-regressors which maximize variance explained by linear readout model alone is
however often unsatisfactory because the constructed readout model may fit into the noise of the response
rather than into the true response curve itself. 
Local regularization appropriately smooths the solution obtained by OFR so that model response does not (over) fit into noise.
Regularization also naturally attenuates instabilities caused by multicollinearity and improves model robustness.

Locally regularized linear readout uses LROFR to estimate $\mathbf{W}^{out}$. The stopping rule~(\ref{eq:LROFR6})
is again left out so that all echo-regressors in $\mathbf{X}$ are analyzed. 
Computation of regularization parameters $\lambda_{i}$ usually converges within 8-10 iterations and
echo-regressors in $\mathbf{X}$ are then ordered based on their importance to the response variable of interest.
The very first iteration can actually be used for observing the unexplained variance ratio of the OFR analysis because
regularization parameters $\lambda_{i}$ are set ,in the beginning, uniformly to extremely small values. 

Observing the evolution of the unexplained variance ratio in the first iteration of LROFR, and
examining vector $\boldsymbol{\lambda}$ with order of regressors in $\mathbf{X}$ give richer and more profound information
concerning the importance of echo-regressors and suitability of an ESN structure with its state-space parameters
in a relation to the unknown dynamics of the underlying problem.

One may change the state-space parameters of an ESN
($\mathbf{W}^{in}$, $\mathbf{W}$, $\mathbf{W}^{fb}$, nonlinearity $\mathbf{f}$, dimensionality, etc.),
carry out LROFR, and observe how different parameters influence the quality and redundancy of echo-regressors.
Such analysis provides in-depth information concerning how an ESN reacts to the data of interest
and may be used for relating particular echo-regressors and overall ESN structure to different features of the training data.
It is important to point out that sole regularization is not able to provide as deep an insight into the efficiency
of an ESN expansion as a joint analysis of variance contributions and Bayesian relevance using LROFR.
This will be experimentally illustrated in Section~\ref{sec:modelingExamples}.

Vector of output weights $\mathbf{W}^{out}$ estimated by LROFR is then re-ordered as was the original order of regressors
in $\mathbf{X}$ for use with the state-space equation.
Alternatively, the unnecessary echo-regressors may be completely left out which
effectively reduces the dimensionality of the readout and is advantageous when further augmenting original echo-regressors by e.g. second power. 
 

\subsection{Locally regularized RBF readout}
Flexibility of a linear readout might sometimes be insufficient for the data under consideration.
Lack of readout flexibility is indicated by an unsatisfactory MSE which indicates that the original
regressors do not explain the response variable sufficiently well and that a transformation of the original
regressors should be carried out. MSE alone, however, does not reveal much information
concerning the entire ESN model. Analysis using LROFR shows how the variance ratio of the error term
(or MSE itself) drops with each selection of a new echo-regressor. 
If the state-space equation generates echo-regressors of low relevance to the response, the unexplained variance ratio
will drop a little with each selection of a new regressor. This indicates that
a more flexible readout will probably explain the response variable with higher accuracy
while requiring less parameters.

Locally regularized RBF readout built using LROFR-DOPT is a flexible nonlinear modeling
strategy that generates parsimonious models with excellent generalization performance.
It is numerically more stable than RBF-RVM and possesses the computational advantages of nonlinear models
linear-in-parameters. Locally regularized RBF readout for an ESN is constructed
as described in Section~\ref{sec:RBF} but vector $\mathbf{x}(k)$ in~(\ref{eq:RBF2}) is combined using
$\mathbf{u}(k), \mathbf{y}(k-1)$ via the state-space equation~(\ref{eq:esnStateSpace});
i.e. $\mathbf{x}(k)$ equals to a $k$-th row of the ESN design matrix $\mathbf{X}$.
The readout may be expressed as
\begin{equation}\label{eq:esnRBFReadout}
\mathbf{y}(k+1) = \mathbf{f}^{out}(\mathbf{W}^{out}(\mathbf{f}^{RBF}(\mathbf{x}(k+1))))
\end{equation}
where $\mathbf{f}^{RBF} = [\phi_{1}(\mathbf{x}(k+1)),\dots,\phi_{M}(\mathbf{x}(k+1))]$ and
$M$ equals to $M_{final}$ after application of LROFR with D-Optimality cost to the full RBF model.

\section{Modeling Examples}
\label{sec:modelingExamples}
The following tasks are used to illustrate the analysis of an ESN model using LROFR algorithms.
Joint orthogonal variance analysis combined with Bayesian relevance in LROFR is experimentally tested. 
Locally regularized RBF readout is additionally used where appropriate
if the analysis indicated a need for a more flexible readout.
The first presented task is a noisy real-world application and the second task is a synthetic noiseless benchmark task. 

\subsection{Synthesis of Handwritten Characters}
An ESN was successfully used to investigate and model the naturalness of handwritten characters
which is defined as being the difference between the strokes of the handwritten characters
and the canonical fonts on which they are based \cite{Dolinsky2009}.
In general, the naturalness learning framework defines naturalness as the difference
between target human-like behavior and the behavior of an original canonical system
which resembles the desired target human-like behavior but lacks human idiosyncrasy.
In the handwriting example, the canonical system comprises the strokes of an original
font character and the naturalness of the differences between handwritten strokes
and their corresponding original font strokes. If it were possible to generate the appropriate naturalness
(differences) for the strokes of a font character, then synthesizing a handwritten character
would be reduced to the simple addition of the naturalness to the original font strokes.

It was shown that the relationship between certain properties
of font strokes (canonical system) and their naturalness exists and can be modeled using an ESN \cite{Dolinsky2009, DolinskyICCC2007}.

Font strokes were expressed as a 2D vector field comprised of vectors between successive evenly spaced points of font strokes.
Naturalness was expressed as a 2D displacement vector field between handwritten strokes and their corresponding original font 
strokes. In other words, an ESN is used to model a relationship between a 2D explanatory variable (font strokes)
and a 2D response (naturalness).
Following the original work \cite{Dolinsky2009}, the training data matrices $\mathbf{U}$ (2D explanatory variable)
and $\mathbf{V}$ (2D response) were established at sizes $2704\times2$. 

The network is comprised of 300 units using the Gaussian RBF activation function with zero mean and variance of $1$. 
The internal weights in the matrix $\textbf{W}$ were randomly assigned values of 0, 0.2073, $-0.2073$ with respective probabilities 0.95, 0.025, 0.025.
For a $300\times300$ matrix $\textbf{W}$, this implies a spectral radius of $\approx0.85$, providing for a relatively long short-term memory \cite{Jaeger2002}.
Two input and two output units were attached. 
Input weights were randomly drawn from a uniform distribution over $(-1,1)$ with probability of 0.9 or set to 0.
With this input matrix, the network is strongly driven by the input activations because many elements of the matrix $\textbf{W}^{in}$ are non-zero values.
The network has output feedback connections $\textbf{W}^{back}$, which were randomly set to one of the three 
values of 0, 0.1, $-0.1$ with respective probabilities 0.9, 0.05, 0.05. 
With this configuration for the feedback connections, the network is only marginally excited by previous output activations; using 
higher values for the feedback connections was found to lead to the network becoming unstable when running on its own \cite{Dolinsky2009}.

The ESN was driven by the samples in $\mathbf{U}$ and $\mathbf{V}$ with a washout period being set to 300. 
The first 300 internal states were discarded. The network internal states $\mathbf{x}(k)$ from $k = 301$ through
$k = 2704$ were collected and saved as rows of the design matrix $\mathbf{X}$.   

\begin{figure}[!ht]
	\centering
	\includegraphics[width=0.5\textwidth]{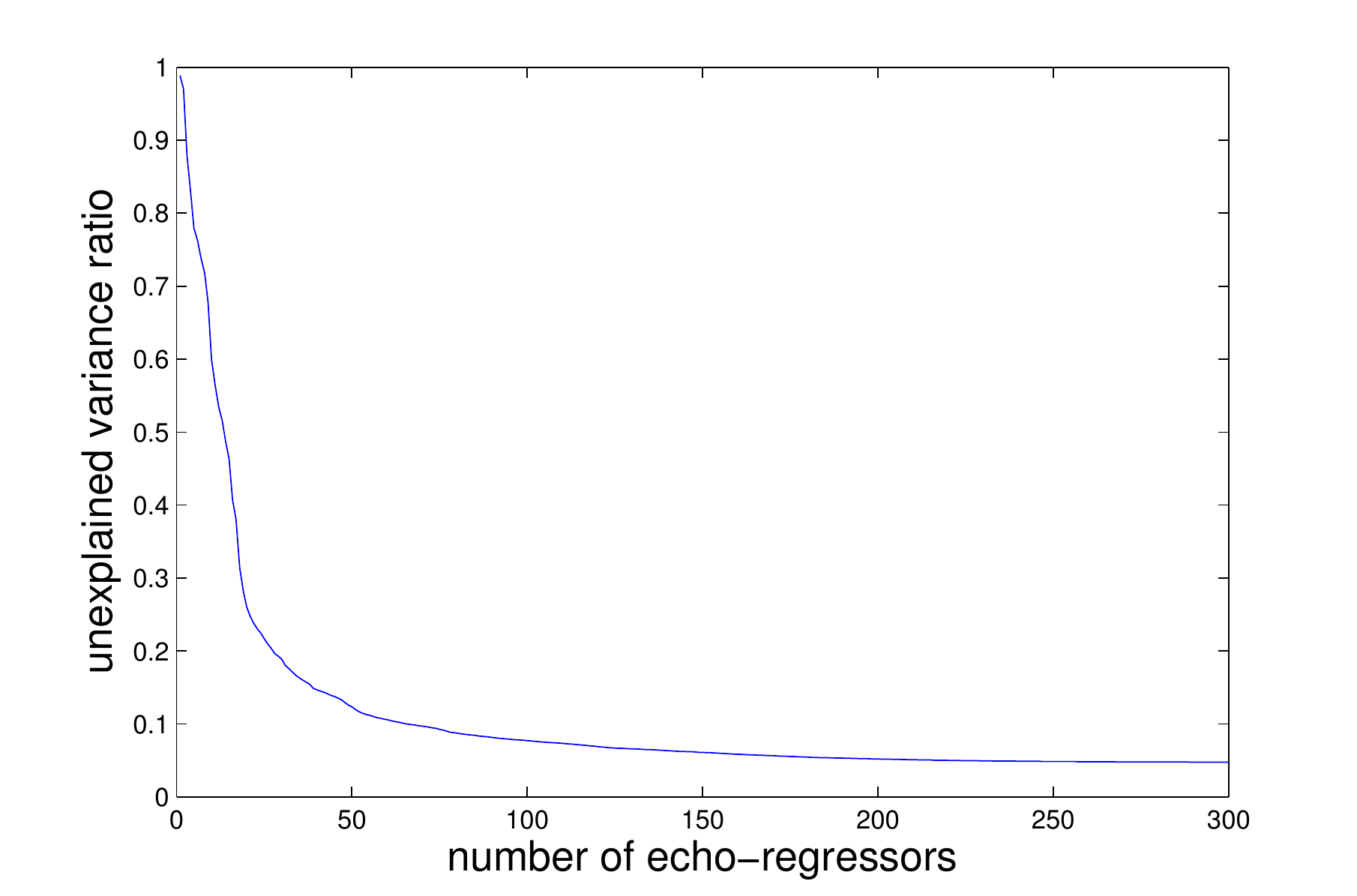}
	\caption{OFR (first iteration of LROFR) using the echo regressors against the $x$ component of the response variable.}
	\label{fig:hwDataOFR1}
\end{figure}

\begin{figure}[!ht]
	\centering
	\includegraphics[width=0.5\textwidth]{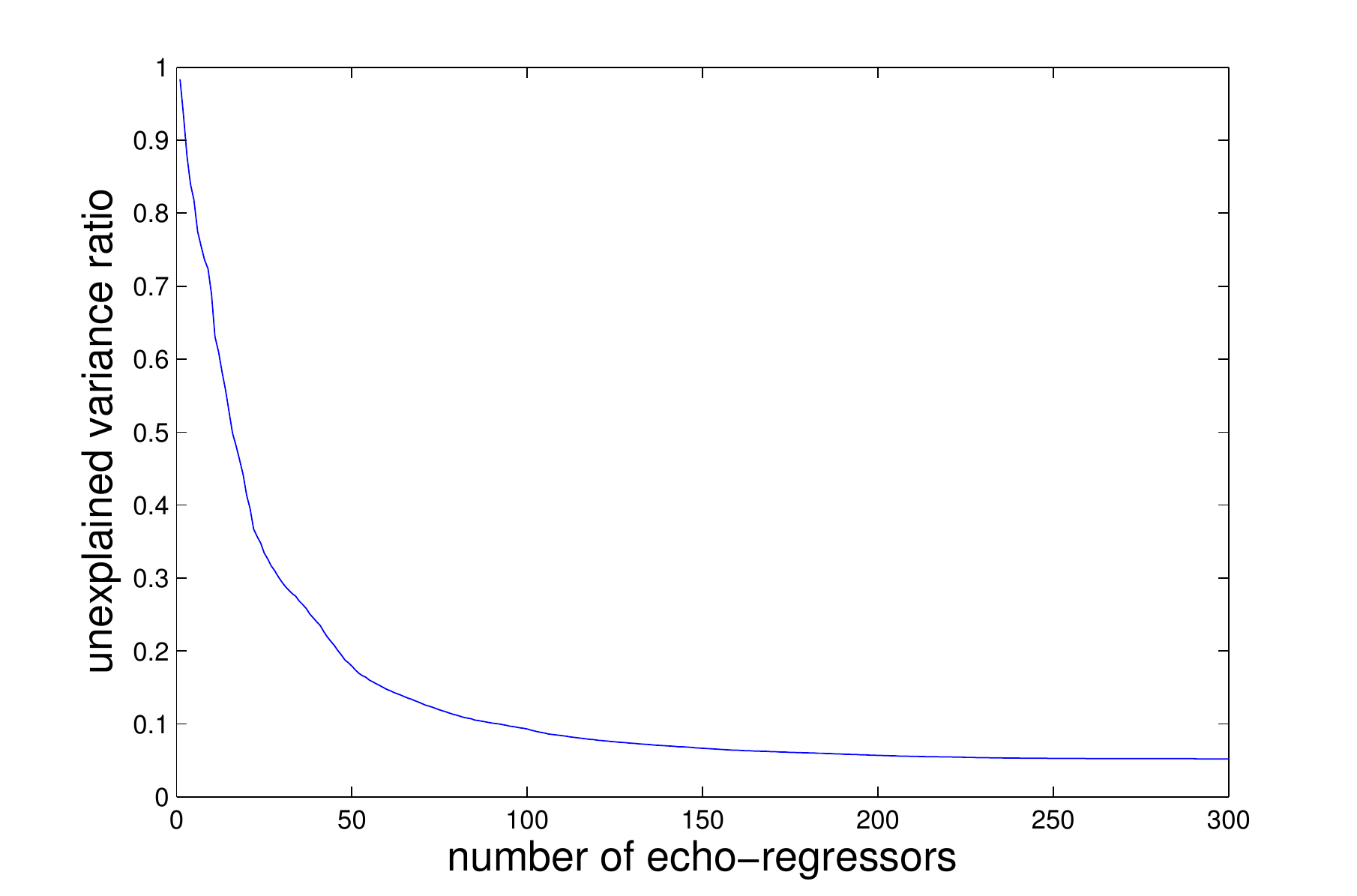}
	\caption{OFR (first iteration of LROFR) using the echo regressors against the $y$ component of the response variable.}
	\label{fig:hwDataOFR2}
\end{figure}

First, the traditional linear readout is used and its performance is observed in the training, testing and modeling stages
\footnote{Training stage uses teacher forcing of the training data. Testing stage uses the same training data but the ESN runs on its own.
Modeling stage uses also inputs (characters) which were not introduced during the training stage and the ESN runs on its own.} (Tab.~\ref{tab:hwData}). 
A relatively low MSE in training is in contrast to a higher MSE in the testing and modeling stages.
Otherwise, not much can be seen from an analysis using sole MSE.
LROFR is then carried out to analyze the statistical quality of the generated echo-regressors in $\mathbf{X}$ against the response variable.
Figures~\ref{fig:hwDataOFR1} and \ref{fig:hwDataOFR2} plot the results of the OFR analysis (first iteration of LROFR).
The unexplained variance ratio (i.e. variance ratio of the error term) decreases sharply and steadily
until about 50 echo-regressors were selected into the readout.
Selecting more echo-regressors provided a moderate but still steady drop in the unexplained variance ratio until about 
150 echo-regressors were selected.
Selecting even more echo-regressors gave a rather negligible drop in the unexplained variance ratio, especially
the region when 180 to 300 echo-regressors were selected. 
Clearly, about half of the echo-regressors contribute very little to the variance of the response variable.

After convergence of the regularization parameters $\lambda_{i}$ (10 iterations), vector $\boldsymbol{\lambda}$
and corresponding regression weights were examined.
About the last 60 echo-regressors had an extremely large $\lambda_{i}$ which in turn caused their corresponding
regression weights to became virtually zero (i.e. extremely small).
The OFR analysis (first iteration of LROFR) however showed that about half (150) of the generated echo-regressors
contribute little to the explained variance of the training response.
This shows that an analysis based purely on regularization (Bayesian relevance) is not able to fully reveal the importance
of the generated echo-regressors. Joint analysis of the individual variance contributions and Bayesian relevance using LROFR gives more
accurate information concerning the importance of the regressors.

The regression weights $\mathbf{W}^{out}$ were then reordered as was the original order of the echo-regressors
in $\mathbf{X}$ so that the readout could be attached to the state-space equation in the testing and modeling stage.
Locally regularizing the linear readout (i.e. attenuating the undesired regressors) increases
the training MSE but improves stability and robustness of the readout, and thus, decreases
the MSE in the testing and modeling stage (see Tab.~\ref{tab:hwData} for details).
 
\begin{table}[ht]
\begin{center}
\caption{Mean square errors and correlation coefficients for all three readouts.
Readout type: \#1 - linear, \#2 - locally regularized linear, \#3 - locally regularized RBF.}
\begin{tabular}{|c|c|c|c|c|}
	\hline
	 {\em readout } & \multicolumn{4}{c|}{{\em training}} \\
	\cline{2-5}
	 {\em type} & \multicolumn{2}{c|}{$mse_{x,y}$} & \multicolumn{2}{c|}{$C_{x,y}$} \\
	\hline
	 {\em \#1} & $9.36\times10^{-4}$	& $6.74\times10^{-4}$	& $0.9758$	& $0.9733$	\\
	\hline
	 {\em \#2} & $9.59\times10^{-4}$	& $6.91\times10^{-4}$	& $0.9752$	& $0.9727$	\\
	\hline
	 {\em \#3} & $1.08\times10^{-3}$	& $1.36\times10^{-3}$	& $0.9719$	& $0.9457$	\\
	\hline
	 & \multicolumn{4}{c|}{{\em testing}} \\
	\cline{2-5}
	 & \multicolumn{2}{c|}{$mse_{x,y}$} & \multicolumn{2}{c|}{$C_{x,y}$} \\
	\hline
	 {\em \#1} & $1.53\times10^{-2}$	& $1.86\times10^{-2}$	& $0.6086$	& $0.3776$	\\
	\hline
	 {\em \#2} & $1.38\times10^{-2}$	& $1.49\times10^{-2}$	& $0.6299$	& $0.4175$	\\
	\hline
	 {\em \#3} & $8.40\times10^{-3}$	& $8.03\times10^{-3}$	& $0.7580$	& $0.6316$	\\
	\hline
	 & \multicolumn{4}{c|}{{\em modeling}} \\
	\cline{2-5}
	 & \multicolumn{2}{c|}{$mse_{x,y}$} & \multicolumn{2}{c|}{$C_{x,y}$} \\
	\hline
	 {\em \#1} & $2.18\times10^{-2}$	& $2.65\times10^{-2}$	& $0.4372$	& $0.2405$	\\
	\hline
	 {\em \#2} & $1.83\times10^{-2}$	& $1.97\times10^{-2}$	& $0.4646$	& $0.2543$	\\
	\hline
	 {\em \#3} & $9.08\times10^{-3}$	& $9.29\times10^{-3}$	& $0.6619$	& $0.4874$	\\
	\hline
\end{tabular}
\label{tab:hwData}
\end{center}
\end{table}

Information that about half of the regressors are of little importance indicates that perhaps a more flexible readout
with less parameters may explain the variance of the response variable with higher accuracy and better robustness.
A locally regularized RBF readout was therefore tried to see whether it could improve performance.
The variance of the Gaussian RBF was set $\upsilon = 65$ and D-Optimality weighting $\beta = 10^{-4}$.
The final RBF readout for the $x$ component of the response had 142 RBFs and for the $y$ component 
of the response 124 RBFs. Results are given in Tab.~\ref{tab:hwData}.
It is evident that the locally regularized RBF readout is superior in the testing and modeling stages.
Interestingly, the number of the RBF nodes is in line with the results of the LROFR analysis given in Fig.~\ref{fig:hwDataOFR1} and Fig.~\ref{fig:hwDataOFR2}.
The figures show that only about 130-150 out of the total 300 regressors are significant.
This suggests that an additional nonlinear flexible readout may explain the response variance with about the same or lower dimensionality
while providing improved accuracy and robustness. Dimensionality of the RBF readouts was found to be similar to the number of significant echo-regressors
and flexibility of the nonlinear RBF response surface resulted in an improved performance as outlined in Tab.~\ref{tab:hwData}.   

The LROFR analysis also suggests that the parameters of the state-space equation in the handwriting
task could be designed and optimized in a better way (e.g. different dimensionality, nonlinearity in $\mathbf{f}$,
weight matrices, etc.) so that less collinear and non-significant echo-regressors are generated.
Although we are here concentrating on an improvement of the readout mechanism itself,
the presented analysis is useful for an improvement and better understanding of an entire ESN model structure
and is discussed further in Section~\ref{sec:discussion}. 

\subsection{Mackey-Glass system}
\begin{figure}[!ht]
	\centering
	\includegraphics[width=0.5\textwidth]{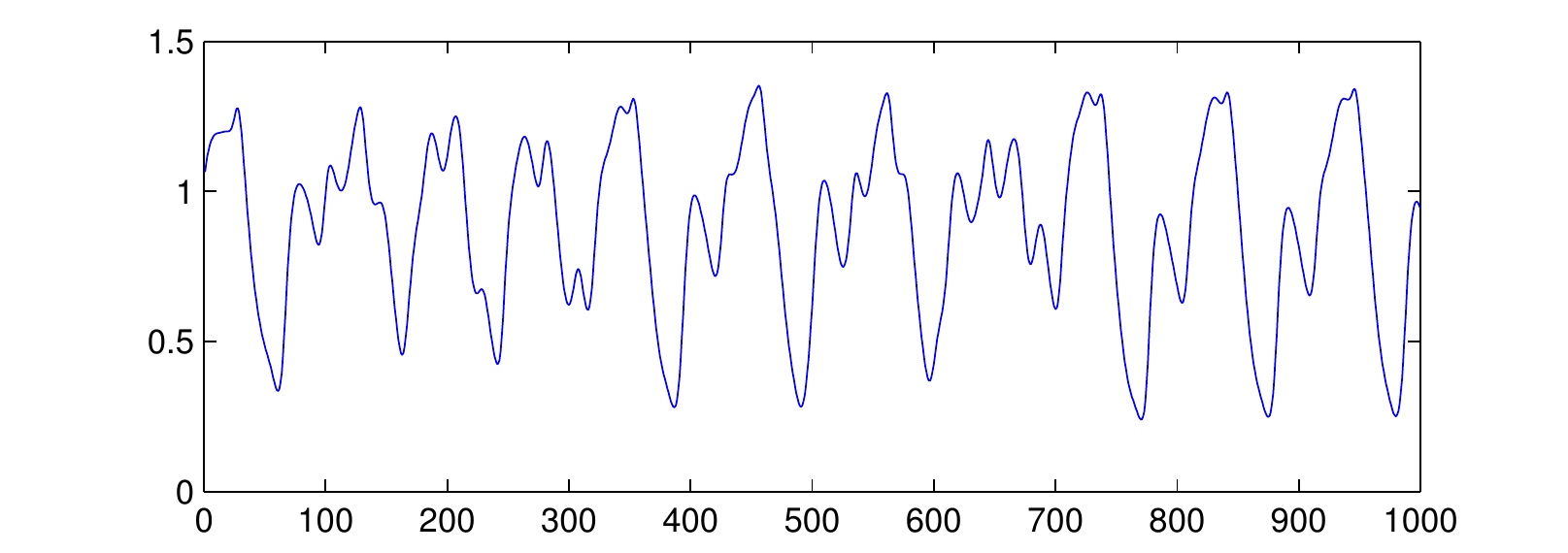}
	\caption{A sequence of the Mackay-Glass system with $\tau = 30$.}
	\label{fig:mg}
\end{figure}

The Mackey-Glass (MG) delay differential equation is a standard benchmark task for time-series modeling.   
\begin{equation}\label{eq:mg1}
\dot{y}(t) = \alpha \frac{y(t-\tau)}{1+y(t-\tau)^{\beta}} - \gamma y(t)
\end{equation}
The system parameters are usually set to $\alpha = 0.2, \beta = 10$, and $\gamma = 0.1$.
If $\tau > 16.8$ then the system has a chaotic attractor. The values of $\tau$ are often set to $17$ (mild chaotic behaviour) or
$30$ (wild chaotic behaviour). In our experiments we set $\tau = 30$. 
We followed the work in \cite{Jaeger2001} and the MG system was approximated in discrete time  
\begin{equation}\label{eq:mg2}
y(k+1) = y(k) + \delta(0.2 \frac{y(k-\tau/\delta)}{1+y(k-\tau/\delta)^{10}} - 0.1 y(k))
\end{equation}
with $\delta$ which denotes a stepsize being set to $\delta=10$ as in \cite{Jaeger2001} so that results may be compared. 
A $3000$ point MG sequence was than generated and used in the training stage.
Figure \ref{fig:mg} depicts the last 1000 points of the training sequence.
The training sequence was transformed $y \to tanh(y-1)$ to fit into interval $(-1,1)$ so that
the sequence may be used with sigmoid internal units.

An ESN used for modeling the MG system comprised of $400$ internal units with standard sigmoid activation function ($tanh$ hyperbolic tangens).  
The internal weights in the matrix $\mathbf{W}$ were randomly assigned values $0, 0.4, -0.4$ with probabilities $0.99, 0.005, 0.005$ respectively.
One input unit was attached to feed in a constant bias of value of $0.2$.
The input weights $\mathbf{W}^{in}$ were randomly set to values $0, 0.14, -0.14$ with respective probabilities $0.5, 0.25, 0.25$.
One linear output unit was attached and the feedback weights $\mathbf{W}^{fb}$ were drawn from a uniform distribution over $[-0.56,0.56]$.
Similarly to \cite{Jaeger2001}, a uniform noise term over $(-10^{-5},10^{-5})$ was included
in the state-space equation to obtain a less correlated echo-regressors and a more stable state-space update.

The training sequence was fed into the network via the feedback weights and the internal states were collected into the design
matrix $\mathbf{X}$. The first 1000 steps were discarded as an initial transient resulting in the design matrix of size of $2000\times400$.
Similarly, the first 1000 points were deleted from the training sequence for the regression task.

The ordinary linear readout was tested first and its MSE was calculated. The MSE was found to be relatively low
which shows that the linear compound of the generated echo-regressors could explain the variance of the training sequence remarkably well.
In fact, the readout gives an almost exact fit to the training sequence.
This is often the case if an ESN setup is strongly autoregressive. In the MG example, the training sequence is fed back
into the reservoir via a relatively large and dense $\mathbf{W}^{fb}$ which in turn causes many of the generated echo-regressors
to strongly correlate with the training sequence (and hence the low MSE in the training stage).

\begin{figure}[!ht]
	\centering
	\includegraphics[width=0.5\textwidth]{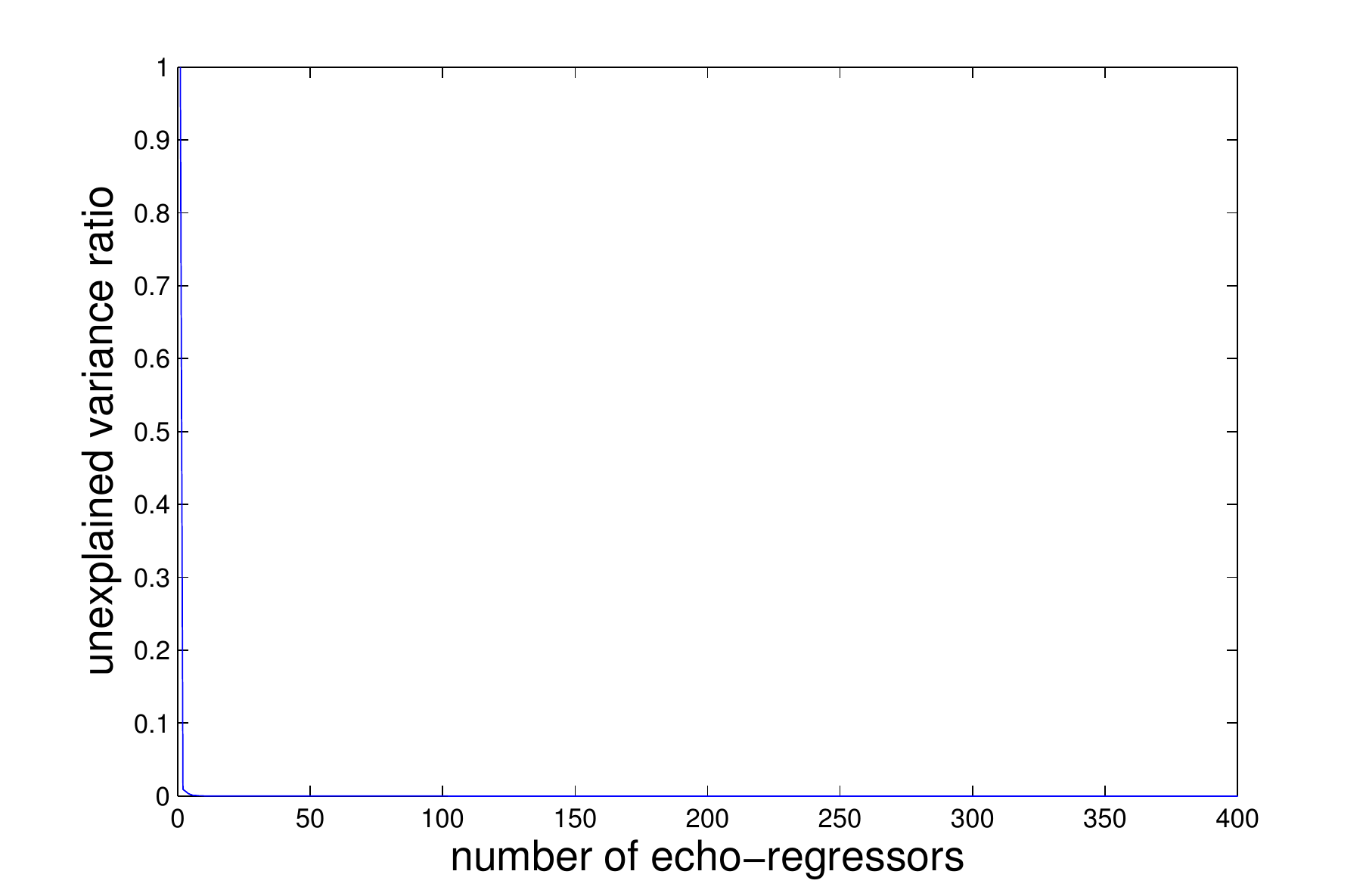}
	\caption{OFR (first iteration of LROFR) using the generated echo-regressors against the response variable.}
	\label{fig:mgOFR}
\end{figure}

The design matrix was further analyzed using OFR.
Figure~\ref{fig:mgOFR} plots the results. It is easy to observe that the unexplained variance sharply drops
with a selection of one single echo-regressor from $\mathbf{X}$.
This can be again attributed to the strongly autoregressive setup of the ESN which generates echo-regressors that are highly
correlated with the training sequence.
This observation indicates that it will probably be difficult to generate a signal of similarly high correlation
using even many RBFs in an RBF readout.
Apparently, in this task a readout capable of generating a flexible response surface will most likely not provide an increase in performance.
This fact together with the autoregressive ESN setup however suggests that delaying the echo-regressors by means of D\&S readout may on the other hand provide an increase in performance.

Locally regularized linear readout was than tested to see how many echo-regressors will be attenuated using regularization parameters.
The analysis was particularly interesting because the training sequence is noise-free.
After the convergence of the regularization parameters $\lambda_{i}$ (10 iterations), vector $\boldsymbol{\lambda}$ 
and the corresponding regression weights were examined.
About 50 echo-regressors had extremely large $\lambda_{i}$ which in turn caused their corresponding
regression weights to became virtually zero (i.e. extremely small).
Table~\ref{tab:mgData} shows the training MSE.
As expected, the locally regularized linear readout gives a slightly higher MSE because its response is smoothed.
Smoothing was ,however, gentle and it was not obvious to the naked eye when compared to the training sequence or to the response
of the ordinary linear readout.
Output weights in $\mathbf{W}^{out}$ were then re-ordered to its original order so that they can be used with the state-space update
in the testing stage.

A locally regularized RBF readout was also tested. The results were however unsatisfactory.
As we already pointed out, this can be attributed to the fact that within the generated echo-regressors there are already signals that
strongly correlate with the response variable. It is difficult and meaningless to generate a signal of similarly high correlation
by adding together even many RBFs (even if each RBF node would have a tunable variance parameter). 

\begin{table}[ht]
\begin{center}
\caption{Mackay-Glass task.
Readout type: \#1 - linear, \#2 - locally regularized linear, \#3 - locally regularized RBF.}
\begin{tabular}{|c|c|c|c|}
	\hline
	 {\em readout type} & $mse_{train}$ & $NRMSE_{84}$ & $NRMSE_{120}$\\
	\hline
	 {\em \#1} & $1.44774\times10^{-8}$	& $0.127$ & $0.219$	\\
	\hline
	 {\em \#2} & $1.51756\times10^{-8}$	& $0.121$ & $0.218$	\\
	\hline
	 {\em \#3} & N/A	& N/A	& N/A \\
	\hline
\end{tabular}
\label{tab:mgData}
\end{center}
\end{table}

In the testing stage, 100 independent sequences of length 1084 were generated. 
All sequences were transformed in the same way as the training sequence ($y \to tanh(y-1)$) to fit them into interval $(-1,1)$ so that
the sequences may be used with a reservoir that has internal units with sigmoid function.
The trained ESN was run teacher-forced for 1000 steps
and then left running free for 84 steps with each of the generated sequences.
The difference between an output of the freely running ESN and a value of the true sequence in step 84 was calculated
and averaged over all 100 trials by means of normalized root mean square error (NRMSE). 
The NRMSE in step 84 is defined using the following formula

\begin{equation}\label{eq:mg3}
NRMSE_{84} = \sqrt{\frac{1}{100\sigma^{2}}\sum_{i=1}^{100}(y_{i}(k+84) - \hat{y}_{i}(k+84))^{2}}
\end{equation}

where $i$ denotes trial number and $\sigma^{2}$ variance of original attractor signal\footnote{We
followed the work in \cite{Jaeger2001} and its source code to be able to compare results.
In fact, the length of the original attractor was $1000 + 84\times100$ instead of $1084\times100$.
After the initial transient period of 1000 points, the internal state of the ESN was saved
and the net was left running free for 84 steps. The difference between the true response and an output of the freely running net
 in step 84 was calculated for its use in the NRMSE formula. Then, the net was ran now teacher-forced from the saved internal state
for 84 steps and new internal state was saved. This procedure was repeated 100 times and NRMSE was computed.}. 
Definition of NRMSE in other steps (e.g. 120) is straightforward.

The two readouts (traditional linear and linear locally regularized) were tested and $NRMSE_{84}$ and $NRMSE_{120}$ were calculated.
Table~\ref{tab:mgData} gives the results. It was interesting to observe that despite the fact that the MG series is noise-free,
a locally regularized readout was found to perform better. It is worth to point out that a locally regularized linear readout
does not give as precise fit to the training MG sequence of 2000 points as the plain linear readout; i.e.
some small fluctuations in the training sequence are smoothed out when using local regularization.
The results in Table~\ref{tab:mgData} suggest that such minor smoothing is beneficial for many (noise-free) sequences in the testing state.

To further investigate this phenomenon we repeated the testing stage ten more times always with new 100 independent sequences.
The results are presented in Table~\ref{tab:mgData2}. As we can see, in some cases a locally regularized linear readout
outperformed a plain linear readout, in some cases, however, it was the linear readout that performed better.
Weights of both readouts were estimated using the MG sequence of 2000 points.
As it is often the case, this training sequence does not capture the entire dynamics of the underlying system and precise fit (linear readout)
works well only if testing sequences are similar to the training one.
For some other sequences, however, the smoothed fit (local regularization) works better.
This is the reason why in Table~\ref{tab:mgData2} there are testing stages where a locally regularized linear readout outperformed
linear readout but also stages where it is the other way around.   
   
\begin{table}[ht]
\begin{center}
\caption{Comparison of $NRMSE_{84}$ and $NRMSE_{120}$ for linear and locally regularized readout in ten independent trials.}
\begin{tabular}{|c|c|c|c|c|}
	\hline
	 	 & \multicolumn{2}{c|}{$NRMSE_{84}$} & \multicolumn{2}{c|}{$NRMSE_{120}$} \\
	\hline
	 {\em readout type} & {\em \#1} & {\em \#2} & {\em \#1} & {\em \#2}	\\
	\hline
	 {\em 1} & $0.108$ & $0.114$ & $0.192$ & $0.202$	\\
	\hline
	 {\em 2} & $0.143$ & $0.139$ & $0.219$ & $0.211$	\\
	\hline
	 {\em 3} & $0.152$ & $0.154$ & $0.273$ & $0.274$	\\
	\hline
	 {\em 4} & $0.165$ & $0.155$ & $0.259$ & $0.242$	\\
	\hline
	 {\em 5} & $0.158$ & $0.163$ & $0.249$ & $0.270$	\\
	\hline
	 {\em 6} & $0.122$ & $0.134$ & $0.154$ & $0.162$	\\
	\hline
	 {\em 7} & $0.125$ & $0.123$ & $0.148$ & $0.147$	\\
	\hline
	 {\em 8} & $0.142$ & $0.141$ & $0.244$ & $0.232$	\\
	\hline
	 {\em 9} & $0.156$ & $0.151$ & $0.246$ & $0.251$	\\
	\hline
	 {\em 10} & $0.156$ & $0.150$ & $0.243$ & $0.241$	\\
	\hline
\end{tabular}
\label{tab:mgData2}
\end{center}
\end{table}

In practice, the length of a training sequence is often limited and even noise-free data may benefit smoothing
a model response using local regularization, especially systems with a periodic attractor. 
Furthermore, LROFR analysis provides
a deeper insight into the dynamics of an ESN model. In the presented MG task, the LROFR analysis (Fig.\ref{fig:mgOFR}) shows that there is no
need for a more flexible readout (e.g. RBF readout) because some of the generated echo-regressors already explain
a high portion of the variance of the training sequence.
Testing trials with both readouts show that there is probably still room for an improvement of performance but
this time not in terms of readout flexibility but rather in terms of extracting more information from the echo-regressors
(e.g. extracting more temporal information using delay\&sum readout \cite{Holzmann2010, ShuEtal2012}).
Some suggestions for research along this line are outlined in the next section.



\section{Discussion}
\label{sec:discussion}
The presented locally regularized readouts improve the accuracy, stability and robustness of an ESN model
either by direct penalization of the undesired echo-regressors using locally regularized linear readout
or by an effective transformation into RBF regressors using a locally regularized RBF readout if more flexibility
is required.

LROFR analysis helps to better understand whether the state-space parameters are suitable for the data under consideration.
Different construction mechanisms for the weight matrices or activation functions give rise to echo-regressors of different shapes and quality.
The presented analysis using the LROFR enables better understanding concerning the efficiency of these mechanisms. 
Some interesting examples of such mechanisms for the LROFR analysis are as follows.
The Hebbian adaptation rule is an interesting mechanism to optimize the weight values in $\mathbf{W}^{in}$, $\mathbf{W}$ ,$\mathbf{W}^{fb}$
for ESN models of moderate size (e.g. 100 internal units) \cite{Babinec2006joei}.
Other works are concerned with adapting the topology of the weight matrices rather than
adapting a value of each particular weight \cite{Babinec2006joei, Deng2007}.        
Imprinting small-world and scale-free structures well known from graph theory onto a structure of the weight matrices
yielded particularly interesting results \cite{Deng2007}. 
Different activation functions $\mathbf{f}$ in the state-space equation has also been studied.
Besides traditional sigmoid or identity functions \cite{Jaeger2001, Jaeger2002}, radial basis functions (RBF) \cite{DolinskyICCC2007},
other non-monotonous functions \cite{Dolinsky2010} and parametrized activation functions in $\mathbf{f}$ \cite{Steil2007}
have been tested with improved performance and stability of the state-space update being observed.
The LROFR analysis is capable of providing profound and detailed insight into these construction mechanisms; regarding their efficiency and suitability
for a task at hand.

Echo-regressors delayed by the D\&S readout may also be analyzed in detail by the LROFR. All the delayed regressors may be analyzed so that the most significant delays are determined.
The importance of the delays may be then used for complex analysis of the temporal structure of the underlying problem.
The analysis will also show whether an additional flexible readout (e.g. RBF) should be used after the D\&S readout so that robustness and accuracy is enhanced even further.   



There is a room for improvements in the presented readouts too.
LROFR may be effectively combined with the PRESS statistics (leave-one-out criterion) so that the locally regularized linear readout
automatically selects a significant subset of echo-regressors with no user-specified stopping rule being involved \cite{Chen2004}. 
This would automatically determine dimensionality for the readout
which is particularly desirable when augmenting internal states.
Another interesting alternative is to use the recently proposed Coordinate descent approach
for finding significant subset of echo-regressors \cite{Friedman2010}.

Generalization abilities and parsimony of locally regularized RBF readout may be also further improved
using tunable RBF kernels \cite{Chen2010pso}.
The outlined research directions are planned to be pursued in the future.

\section{Conclusion}
\label{sec:conclusion}
This study has shown that the presented modeling strategy enables better understanding, design and evaluation of ESN models.
Both presented readouts improve the generalization ability of an ESN and are viable
alternatives to traditional linear readout or nonlinear readouts based on FFNN/RBF-RVM.
Improved performance is not the only advantage of the presented approach.
The importance of echo-regressors generated via the ESN state-space equation can be transparently
analyzed using the presented strategy. Suitability of the state-space parameters and overall ESN structure for a task at hand
can therefore be inspected in a straightforward manner.

Future work will be targeted towards the improvements of our strategy which are outlined in Section~\ref{sec:discussion}.
The proposed improvements are likely to further advance the design of ESN models producing stable and parsimonious
ESN models that generalize well.

\section*{Acknowledgement}
This study was supported in part by Japanese Society for Promotion of Science (JSPS) Grant-in-Aid No. 21-09702.

\bibliographystyle{IEEEtran}
\bibliography{biblio.bib}

\begin{thebibliography}{10}
\providecommand{\url}[1]{#1}
\csname url@rmstyle\endcsname
\providecommand{\newblock}{\relax}
\providecommand{\bibinfo}[2]{#2}
\providecommand\BIBentrySTDinterwordspacing{\spaceskip=0pt\relax}
\providecommand\BIBentryALTinterwordstretchfactor{4}
\providecommand\BIBentryALTinterwordspacing{\spaceskip=\fontdimen2\font plus
\BIBentryALTinterwordstretchfactor\fontdimen3\font minus
  \fontdimen4\font\relax}
\providecommand\BIBforeignlanguage[2]{{%
\expandafter\ifx\csname l@#1\endcsname\relax
\typeout{** WARNING: IEEEtran.bst: No hyphenation pattern has been}%
\typeout{** loaded for the language `#1'. Using the pattern for}%
\typeout{** the default language instead.}%
\else
\language=\csname l@#1\endcsname
\fi
#2}}

\bibitem{Jaeger2001}
H.~Jaeger, ``The `echo-state' approach to analysing and training recurrent
  neural networks,'' Fraunhofer Institute for Autonomous Intelligent Systems,
  GDM Report 148, December 2001.

\bibitem{Jaeger2002}
------, ``Tutorial on training recurrent neural networks, covering bppt,rtrl,
  ekf and the {echo state network} approach.'' Fraunhofer Institute for
  Autonomous Intelligent Systems, GDM Report 159, October 2002.

\bibitem{Ishii2004}
K.~Ishii, T.~van~der Zant, V.~Becanovi\'c, and P.~Ploger, ``Identification of
  motion with echo state network,'' in \emph{Oceans'04}.\hskip 1em plus 0.5em
  minus 0.4em\relax IEEE, November 2004.

\bibitem{Babinec2006joei}
{\v{S}}.~Babinec, ``Optimalization echo state neural networks with hebbian
  learning and genetic algorithm,'' \emph{Journal of Electrical Engineering},
  vol.~56, no.~12, pp. 28--31, 2006.

\bibitem{Deng2007}
Z.~Deng and Y.~Zhang, ``Collective behavior of a small-world recurrent neural
  system with scale-free distribution,'' \emph{IEEE Transactions on Neural
  Networks}, vol.~18, no.~5, pp. 1364--1375, 2007.

\bibitem{Steil2007}
J.~J. Steil, ``Online reservoir adaptation by intrinsic plasticity for
  backpropagation-decorrelation and echo state learning.'' \emph{Neural
  Networks}, vol.~20, no.~3, pp. 353--364, 2007.

\bibitem{Dolinsky2010}
J.~Dolinsk\'y and S.~Konishi, ``Echo-state networks with non-monotonous
  activation functions,'' in \emph{The 2010 Japanese Joint Statistical
  Meeting}.\hskip 1em plus 0.5em minus 0.4em\relax Tokyo, Japan: JSS, September
  2010.

\bibitem{Babinec2006icann}
{\v{S}}.~Babinec and J.~Posp\'ichal, ``Merging echo state and feedforward
  neural networks for time series forecasting,'' in \emph{Artificial Neural
  Networks – ICANN 2006}, ser. LNCS 4131.\hskip 1em plus 0.5em minus
  0.4em\relax Springer, 2006, pp. 367--375.

\bibitem{Holzmann2010}
\BIBentryALTinterwordspacing
G.~Holzmann and H.~Hauser, ``Echo state networks with filter neurons and a
  delay\&sum readout,'' \emph{Neural Networks}, vol.~23, no.~2, pp. 244 -- 256,
  2010. [Online]. Available:
  \url{http://www.sciencedirect.com/science/article/pii/S0893608009001580}
\BIBentrySTDinterwordspacing

\bibitem{Ozturk2007}
\BIBentryALTinterwordspacing
M.~C. Ozturk and J.~C. Principe, ``An associative memory readout for esns with
  applications to dynamical pattern recognition,'' \emph{Neural Networks},
  vol.~20, no.~3, pp. 377 -- 390, 2007, echo State Networks and Liquid State
  Machines. [Online]. Available:
  \url{http://www.sciencedirect.com/science/article/pii/S0893608007000329}
\BIBentrySTDinterwordspacing

\bibitem{Dutoit2009}
\BIBentryALTinterwordspacing
X.~Dutoit, B.~Schrauwen, J.~V. Campenhout, D.~Stroobandt, H.~V. Brussel, and
  M.~Nuttin, ``Pruning and regularization in reservoir computing,''
  \emph{Neurocomputing}, vol.~72, no. 7 E, pp. 1534 -- 1546, 2009,
  <ce:title>Advances in Machine Learning and Computational
  Intelligence</ce:title> <ce:subtitle>16th European Symposium on Artificial
  Neural Networks 2008</ce:subtitle> <xocs:full-name>16th European Symposium on
  Artificial Neural Networks 2008</xocs:full-name>. [Online]. Available:
  \url{http://www.sciencedirect.com/science/article/pii/S0925231209000186}
\BIBentrySTDinterwordspacing

\bibitem{ShuEtal2012}
\BIBentryALTinterwordspacing
D.~Shutin, C.~Zechner, R.~Kulkarni, and H.~Poor, ``{Regularized Variational
  Bayesian Learning of Echo State Networks with Delay\&Sum Readout},''
  \emph{Neural Computation}, vol.~24, no.~4, pp. 967--995, 2012. [Online].
  Available:
  \url{http://control.ee.ethz.ch/index.cgi?page=publications;action=details;id%
=3909}
\BIBentrySTDinterwordspacing

\bibitem{KroseSmagt96}
\BIBentryALTinterwordspacing
B.~Krose and P.~van~der Smagt, ``An introduction to neural networks,'' 1996.
  [Online]. Available:
  \url{citeseer.ist.psu.edu/article/krose93introduction.html}
\BIBentrySTDinterwordspacing

\bibitem{Tipping2001}
M.~E. Tipping, ``Sparse bayesian learning and the relevance vector machine,''
  \emph{Journal of Machine Learning Research}, vol.~1, pp. 211--244, 2001.

\bibitem{Chen2006}
S.~Chen, ``Local regularization assisted orthogonal least squares regression,''
  \emph{Neurocomputing}, vol.~69, pp. 559--585, 2006.

\bibitem{Chen1989}
S.~Chen, S.~A. Billings, and W.~Luo, ``Orthogonal least squares methods and
  their application to non-linear system identification,'' \emph{International
  Journal of Control}, vol.~50, no.~5, pp. 1873--1896, 1989.

\bibitem{Wong1935}
Y.~K. Wong, ``An application of orthogonalization process to the theory of
  least squares,'' \emph{Annals of Mathematical Statistics}, vol.~6, no.~2, pp.
  53--75, 1935.

\bibitem{Chen2003}
S.~Chen, X.~Hong, and C.~J. Harris, ``Sparse kernel regression modeling using
  combined locally regularized orthogonal least squares and d-optimality
  experimental design,'' \emph{IEEE Transactions on Automatic Control},
  vol.~48, no.~6, pp. 1029--1036, 2003.

\bibitem{Dolinsky2009}
J.~Dolinsk\'y and H.~Takagi, ``Analysis and modeling of naturalness in
  handwritten characters,'' \emph{IEEE Trans. on Neural Networks}, vol.~20,
  no.~10, pp. 1540--1553, 2009.

\bibitem{DolinskyICCC2007}
------, ``Synthesizing handwritten characters using naturalness learning,'' in
  \emph{Int. Conf. on Computational Cybernetics (ICCC2007)}.\hskip 1em plus
  0.5em minus 0.4em\relax Gammarth, Tunisia: IEEE, August 2007, pp. 101--106.

\bibitem{Chen2004}
S.~Chen, X.~Hong, C.~J. Harris, and P.~M. Sharkey, ``Sparse modeling using
  orthogonal forward regression with press statistic and regularization,''
  \emph{IEEE Transaction On Systems, Man, and Cybernetics-Part B: Cybernetics},
  vol.~34, no.~2, pp. 898--911, 2004.

\bibitem{Friedman2010}
\BIBentryALTinterwordspacing
J.~H. Friedman, T.~Hastie, and R.~Tibshirani, ``Regularization paths for
  generalized linear models via coordinate descent,'' \emph{Journal of
  Statistical Software}, vol.~33, no.~1, pp. 1--22, 2 2010. [Online].
  Available: \url{http://www.jstatsoft.org/v33/i01}
\BIBentrySTDinterwordspacing

\bibitem{Chen2010pso}
S.~Chen, X.~Hong, and C.~J. Harris, ``Particle swarm optimization aided
  orthogonal forward regression for unified data modeling,'' \emph{IEEE
  Transaction On Evolutionary Computation}, vol.~14, no.~4, pp. 477--499, 2010.

\end{thebibliography}

\end{document}